\documentclass{article}

\usepackage{PRIMEarxiv}

\usepackage[utf8]{inputenc} 
\usepackage[T1]{fontenc}    
\usepackage{hyperref}       
\usepackage{url}            
\usepackage{booktabs}       
\usepackage{amsfonts}       
\usepackage{nicefrac}       
\usepackage{microtype}      
\usepackage{lipsum}
\usepackage{fancyhdr}       
\usepackage{changepage}
\usepackage{graphicx}       
\usepackage{tabularx}       
\usepackage{multirow}
\usepackage{caption}
\usepackage{siunitx}
\usepackage{placeins}
\usepackage{subfig}

\title{Joint object detection and re-identification for 3D obstacle multi-camera systems
}

\author{
  Irene Cortés \\
  Department of Systems Engineering and Automation \\
  Universidad Carlos III de Madrid  \\
  Leganés\\
  \texttt{irecorte@ing.uc3m.es} \\
   \And
  Jorge Beltrán \\
  Department of Signal Theory, Telematics, and Computer Science \\
  Universidad Rey Juan Carlos  \\
  Fuenlabrada\\
\And
  Arturo de la Escalera \\
  Department of Systems Engineering and Automation \\
  Universidad Carlos III de Madrid  \\
  Leganés\\
  \And
  Fernando García \\
  Department of Systems Engineering and Automation \\
  Universidad Carlos III de Madrid  \\
  Leganés\\
}

\begin{document}
\maketitle

\begin{abstract}
In recent years, the field of autonomous driving has witnessed remarkable advancements, driven by the integration of a multitude of sensors, including cameras and LiDAR systems, in different prototypes. However, with the proliferation of sensor data comes the pressing need for more sophisticated information processing techniques. This research paper introduces a novel modification to an object detection network that uses camera and lidar information, incorporating an additional branch designed for the task of re-identifying objects across adjacent cameras within the same vehicle while elevating the quality of the baseline 3D object detection outcomes.
The proposed methodology employs a two-step detection pipeline: initially, an object detection network is employed, followed by a 3D box estimator that operates on the filtered point cloud generated from the network's detections. Extensive experimental evaluations encompassing both 2D and 3D domains validate the effectiveness of the proposed approach and the results underscore the superiority of this method over traditional Non-Maximum Suppression (NMS) techniques, with an improvement of more than 5\% in the car category in the overlapping areas. 
\end{abstract}


\section{INTRODUCTION}
In the previous decade, research on perception systems for autonomous driving was mainly focused on understanding the traffic situation in front of the vehicle, mainly powered by popular datasets such as KITTI~\cite{kitti_object} or Cityscapes~\cite{cityscapes}. Although this approach limits the complexity of the problem and the computational requirements, its outcome restricts the set of use cases to simple scenarios where an automated operation of the car can be safely performed (e.g. highways).

More recently, the increase in GPU capabilities and the great advances in deep learning models have led to the emergence of new research platforms and prototypes targeted to drive in more challenging traffic environments, with a higher degree of interaction with other road users and involving difficult maneuvers, like in cities \cite{nuscenes, argoverse}. In order to achieve this milestone, the perception pipeline of an Autonomous Vehicle must be able to identify the different participants and potential hazards in the whole scene, not only in the forward direction.

To keep pace with these new needs, more complex sensor configurations aimed at covering the whole horizontal field of view around the vehicle have become the prevailing trend. Nowadays, using setups composed of multiple cameras and one or more LiDAR devices is a common choice to capture meaningful information in 360\si{\degree} with sensor redundancy, so that safe navigation can be achieved \cite{janai2020computer}.

Apart from redundancy, the popularization of perception systems made of multiple units of different technologies brings many opportunities to build robust pipelines capable of providing a precise understanding of the driving environment. For example, combining the data from cameras and LiDAR devices allows for a better estimation of object positions, sizes, and velocities. Furthermore, fusing heterogeneous information from different modalities can help overcome each sensor's limitations, such as the incapability of cameras to perceive objects in low light situations or the sensitivity of LiDARs to adverse weather conditions like rain or fog.

However, all these advances also bring with them new challenges that must be addressed, such as extrinsic calibrations between sensors of different types \cite{beltran2022automatic}, the synchronization of all sensors to obtain information at known time instants \cite{milanes2021tornado}, and the merging of data and detections between sensors of the same and different types \cite{kinzig2022real, cortes2020sianms}. Nevertheless, the potential benefits of these perception systems for autonomous driving, including increased safety and efficiency promising.

In this paper, we propose a solution to address the last mentioned challenge, which manages the entire space around the vehicle by merging detections in areas covered by more than one camera of the same type and would otherwise be detected in duplicate or with partial information. 
The work is based on a two-step 3D detection pipeline \cite{beltran2020towards}, which uses a 2D detection model in the camera modality to generate candidates so that the LiDAR cloud can be filtered into smaller regions of interest to estimate final 3D boxes efficiently. While the performance of this approach was generally robust in a variety of traffic situations, the architecture suffered when objects appeared on the overlapping areas of consecutive cameras due to truncated or duplicate detections.

To tackle this pitfall, the image detection network is modified to include the re-identification module presented in \cite{cortes2020sianms} as a third branch. 
In the second step, the point cloud subsets associated with those detections identified as belonging to the same obstacle are merged. That way, the filtered 3D information for each obstacle does not suffer truncations due to limits in the horizontal field of view (HFOV) of the cameras.
As a result, the proposed pipeline ensures that the estimation of the parameters of every obstacle is based on its complete representation in the 3D space, avoiding the inference of a single instance from multiple partial views.

The remainder of this paper is organized as follows. In Section~\ref{sec:related}, a brief review of related work is provided. Section~\ref{sec:approach} presents a general overview of the proposed algorithm. Section~\ref{sec:experiments} provides experimental results that assess the performance of the method. Finally, conclusions and open issues are discussed in Section~\ref{sec:conclusion}. 
\global\csname @topnum\endcsname 0
\global\csname @botnum\endcsname 0

\section{RELATED WORK}
\label{sec:related}

Several significant advancements have been noted in recent years in the field of 3D obstacle detection with multi-camera systems. The first topic examines the evolution of in-vehicle object detection datasets. Compared to older datasets like KITTI, newer ones provide detailed annotations under diverse conditions. The next area tackles the challenge of managing multiple detections of a single object. Several techniques, from improved Non-Maximal Suppression methods to the innovative Deterministic Point Process, have been developed to address this. The final section explores Siamese networks' role in re-identifying obstacles, a vital task when dealing with views from multiple cameras. These networks have shown their value by distinguishing small differences among objects in crowded settings.

\subsection{In-Vehicle Object Detection Datasets}
The pursuit for greater automation levels in the automotive industry has led to a demand for richer annotated datasets that allow perception systems to cope with more complex traffic scenarios.

Compared to the classic KITTI benchmark, with daytime-only frames and focused on the forward direction, recent datasets have multiple cameras and LiDARs, recorded in both day and night situations, in rain or fog:
Datasets like Waymo \cite{waymo} consist of data from 5 LiDARs and 5 cameras, annotated both in 2D and 3D. Argoverse \cite{argoverse} features two LiDARs and seven cameras with 3D box annotations. PandaSet \cite{pandaset} combines data from six cameras and two LiDARs, providing annotations per LiDAR point and 3D boxes. Similarly, the nuScenes dataset \cite{nuscenes} integrates one LiDAR, six cameras, and multiple radars, offering annotated 3D boxes for each object.
These datasets reflect a growing trend toward more complex and information-rich perception systems, leveraging multiple sensors to capture a more complex, complete, and detailed view of the surrounding environment. However, with the inclusion of multiple cameras comes the inherent challenge of efficiently handling and processing redundant and sometimes conflicting information between views. Duplicate detections, occlusions, and truncation of objects at the edges of the field of view are common problems that arise in multi-camera configurations. In addition, variability in lighting and environmental conditions, such as night, rain, or fog, adds another layer of complexity to data processing and analysis. 

\subsection{Elimination of Multiple Detections for the Same Element}
Several papers have explored techniques related to reducing the number of detections for the same element. The suppression of redundant detections is essential to ensure accurate and consistent perception.
In \cite{bodla2017soft}, a variant of the traditional Non-Maximal Suppression (NMS) method called Soft-NMS is proposed. Unlike traditional NMS which discards detections based on a predefined threshold, Soft-NMS modifies detection scores continuously, resulting in improved detection accuracy. 
In crowded environments, pedestrian detection can be challenging due to multiple overlapping detections. In \cite{liu2019adaptive}, this problem is addressed by dynamically adapting the NMS threshold according to the density of detections in the local region, enabling better discrimination between nearby pedestrians.
In \cite{some2020determinantal}, the use of Deterministic Point Process (DPP) is introduced as an alternative to traditional NMS. DPP selects a subset of detections that are representative and diverse, which can be beneficial in crowded scenarios where detections overlap.
Finally, \cite{liu2019learning} focuses on learning pairwise relationships between detections to improve accuracy in crowded scenes. The method can reduce false detections and improve discrimination between nearby objects by explicitly modeling these relationships.


\subsection{Siamese Networks for Obstacle Re-Identification}
On multi-camera systems, the management of multiple detections of a single object needs to be handled in a different fashion, as NMS techniques cannot be applied to bounding boxes belonging to images captured from different perspectives (e.g. distributed along the roof of the car). Even so, there are usually parts of the environment that are seen from two cameras, due to an overlap between their horizontal fields of view. For specific situations where the cameras are in close proximity to each other, methods such as image-stitching can be used to form a large panoramic image composed of images from several cameras \cite{kinzig2022real, xiang2016image, lin2015cylindrical}. In a more general approach, this problem is often solved through the use of agent re-identification networks, which allow to get rid of redundant detections based on feature similarities \cite{cortes2020sianms, he2021transreid}. 

Siamese networks are a powerful tool in object re-identification, especially in scenarios with multiple obstacles and crowded environments. 
In \cite{koch2015siamese}, Siamese networks are used to learn to recognize objects from a single example by leveraging the network structure to compare images. This capability is essential in scenarios where data collection is costly or impractical.
Another example is \cite{liu2016deep}, where this type of network is applied to differentiate between similar vehicles. By learning relative distances, the network is able to identify subtle differences between vehicles that at first glance appear identical. This technique is especially useful in applications such as urban surveillance, where accurate vehicle re-identification is essential.
Examples of the use of Siamese networks can be seen in \cite{zheng2017person, yi2014deep} or in \cite{gomez2018deep}, where re-identification is approached from a "parts" perspective: instead of treating the person as a whole, the network is trained to recognize and compare individual parts (such as head, torso, legs), allowing for greater robustness against varied occlusions and postures.

\begin{figure}[!ht]
\centering
\includegraphics[width=\textwidth]{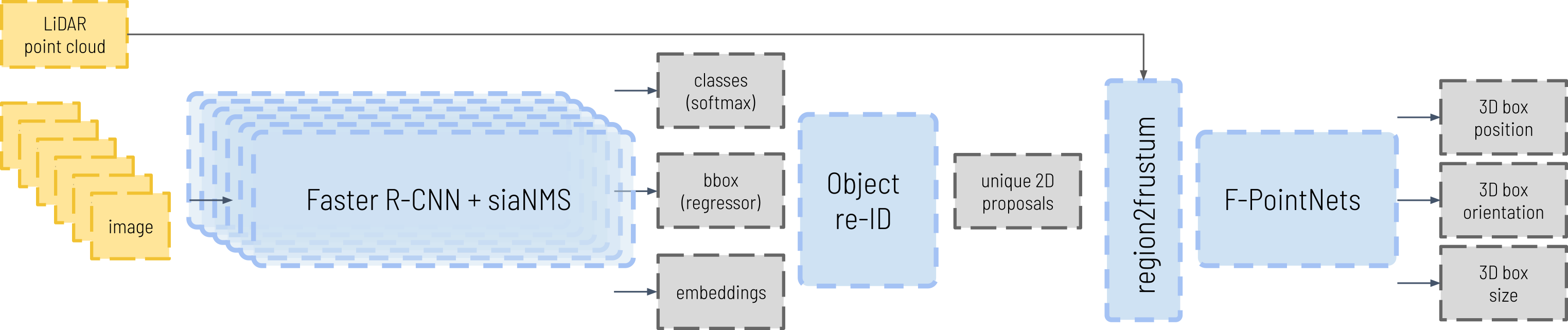}
\caption{Complete detection pipeline}
\label{fig:global_graph}
\end{figure}

\section{PROPOSED APPROACH}
\label{sec:approach}

In this paper, we attempt to address the issues identified in the two-step multicamera-lidar detection system presented in \cite{beltran2020towards} by endowing the image detector with a new re-identification branch to better handle duplicate or truncated detections by the HFOV of the cameras. 

In the selected pipeline a 3D obstacle detection is performed in two steps; the first one consists of a generation of image proposals, obtained with the Faster R-CNN network \cite{ren2015faster, wu2019detectron2} in each of the vehicle's cameras. 

Subsequently, by means of the extrinsic calibration parameters, a set of frustums will be obtained by filtering the LiDAR point cloud with each of the detected bounding boxes, which will be used as input to the second step, the Frustum PointNets network \cite{qi2018frustum}, obtaining a 3D box for each of the proposed obstacles. A complete outlook of the whole pipeline can be seen in Figure~\ref{fig:global_graph}.

Ideally, all the proposals would be unique for each obstacle and would cover it completely, surrounding the whole area of the object visible from the ego-car. Unfortunately, this is not feasible, partly because of the possible failures of the detector in the image, but also due to the limited horizontal field of view (HFOV) of the cameras, which will generate truncations in the detections and possible duplicates in obstacle detections that appear in several contiguous cameras at the same time.

To tackle this, this work integrates the siaNMS module presented in \cite{cortes2020sianms}, allowing the end-to-end training of the model. In addition, incorporating the module as a branch of the original network ensures that the learned feature map encodes meaningful characteristics for the three tasks: class prediction, bounding box regression, and re-identification. Furthermore, utilizing this information when filtering the point cloud facilitates obtaining a single frustum per obstacle by combining the matched regions of interest from both images. 

\subsection{Re-Identification Branch Description}
The objective of the re-identification branch introduced in the detection network is to obtain an embedding that encodes the information obtained from each obstacle. The network is trained in such a way that the embeddings generated for the different detections of the same object are very similar 
while maximizing the differences between detections of distinct objects.

For this purpose, we start from a Faster R-CNN detection network structure, composed of a convolutional backbone, from which regions of interest are obtained by means of a Region Proposal Network (RPN) stage. The encoded tensor is cut using these candidates and an ROI Align layer is used to obtain a feature map of constant size for each proposal. These maps are fed to the Fully Connected (FC) layers stage, from which we obtain the outputs of the network: in the case of the original network that was the class of the object and its bounding box.

\begin{figure}[!t]
\centering
\includegraphics[width=\textwidth]{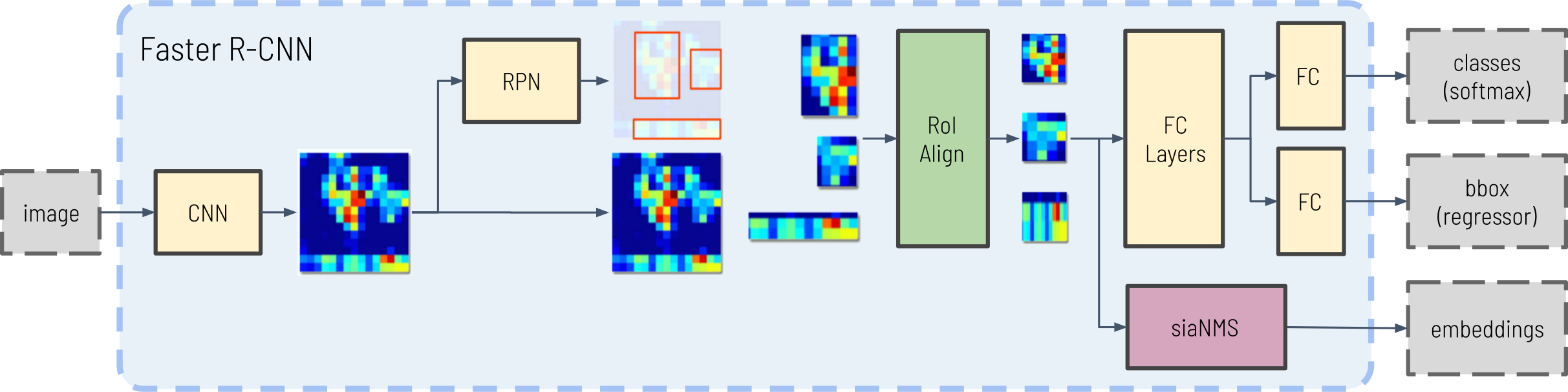}
\caption{Image detection network}
\label{fig:sianms_net}
\end{figure}

In this paper, a third output branch is added tailored for re-identification. This branch is also fed with the fixed-size feature maps obtained for each network proposal and consists of a series of convolutional and FC layers that compress the information of each obstacle into an embedding, with fixed output dimensions, \textit{d}. A schematic of the modified network structure can be seen in Figure~\ref{fig:sianms_net}.

\subsection{Training Data Organization}
In the original 2D stage, a set of annotated images is passed as input in a randomized order through the network (grouped in batches) until the entire data set (epoch) is completed. This is repeated as many times as necessary until the network is sufficiently trained. 
Although this is the standard procedure for conventional image detectors, this way of ordering the data is unsuitable for the purpose of this paper since we need the same object to appear several times in the same forward pass of the network so that the re-identification branch can be trained. 

Intuitively, in order to train the class and bounding box branches we need the annotations of the obstacles present in each image. However, to train the re-identification branch the type of annotation needed is different, namely which pairs of annotations correspond to the same obstacle and which do not. Therefore, in each batch of images, we need examples of both negative pairs of detections, which do not correspond to the same object, and positive ones, which are detections of the same object. 

In this study, we have used the nuScenes dataset, composed of 6 cameras distributed in such a way that they cover 360º around the car. 
Taking advantage of this, we have trained the network using batches of 6 images, one corresponding to each of the surrounding cameras. This way, the input of the model during the training process matches the one that will be used in the inference, while objects appearing in regions shared by contiguous cameras can serve as positive pairs for the training. 
That said, the order of time instants for the 6-image batches is randomized, mimicking the behavior of traditional CNN training.

\subsection{Network Training and Loss Function Definition}
During training, the process explained below is performed for each batch of images: First, a forward pass is made, and with the results obtained at the output of the class and box branches, a loss is calculated for each image individually:

\begin{equation}
\mathcal{L}_{box\_head} = \sum_{i=0}^{F}\mathcal{L}_{box\_reg, i} + \sum_{i=0}^{F+B}\mathcal{L}_{cls, i},
\end{equation}
where $F$ and $B$ are the numbers of foreground and background detections in that image, $\mathcal{L}_{box\_reg, i}$ is the Smooth-L1 Loss, and $\mathcal{L}_{cls, i}$ is the Cross-Entropy Loss.

Once the loss of the detections of the batch images has been acquired, the loss of re-identification between the detections of the 6 images is calculated. The total loss of the batch is the sum of all the losses of the individual images plus the re-identification loss:
\begin{equation}
\mathcal{L}_{batch} = \mathcal{L}_{reID} + \sum_{n=0}^{N}\mathcal{L}_{box\_head, n}, 
\end{equation}
where N is the number of images in the batch, 6 for this case. This way, the gradients calculated will take into account the information on the three kinds of output generated by the network.  

To calculate the re-identification loss, all the possible pairs between the foreground detections (those with an IoU > 0.7 with the Ground Truth) are obtained, that is, all the possible combinations between the detections of the batch images. The loss will be the sum of the individual losses of all positive pairs and the same number of negative pairs. To select which negative pairs are used, their losses are sorted in decreasing order and the ones with the highest loss value are chosen, following an Online Hard Example Mining technique (OHEM), as presented in \cite{shrivastava2016ohem}, 

Then, the Double Margin Contrastive Loss \cite{gomez2018deep} is calculated for each pair, as explained in \cite{cortes2020sianms}: 

\begin{equation}
\label{eq:triplet_loss}
        \mathcal{L}_{reID} = 
        \frac{1}{2} \sum_{i}^{N}\Big[\max \left( \left\|f\left(x_{i}^{r}\right)-f\left(x_{i}^{p}\right)\right\|_{2}-\alpha, 0 \right)^2 + \max \left(\beta - \left\|f\left(x_{i}^{r}\right)-f\left(x_{i}^{n}\right)\right\|_{2}, 0 \right)^2\Big],
\end{equation}
where $\alpha$ and $\beta$ are the two constant margins and $f\left(x_{i}^{r}\right)$, $f\left(x_{i}^{p}\right)$ and $f\left(x_{i}^{n}\right)$ are the embeddings of the reference object, their positive pair and the hardest negative pair, respectively. 

\subsection{Re-Identification evaluation}
 \label{sec:reid-eval}
\begin{figure}[ht]
\centering
\captionsetup[subfloat]{justification=centering}
\subfloat[The six images of the same instant, with the detections obtained in each image.]{
\includegraphics[width=0.975\linewidth]{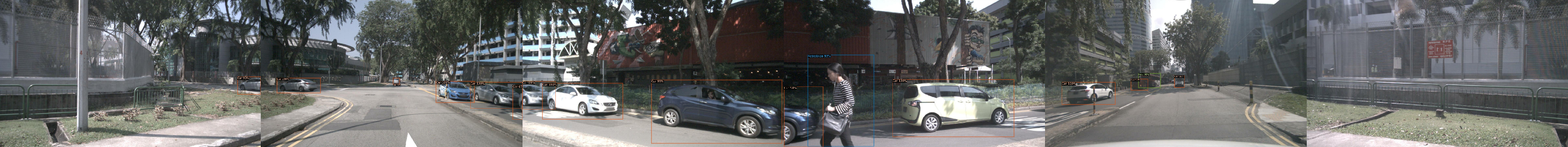}
} \\
\subfloat[Obtained distance matrix between the detections of the front and right cameras]{
\includegraphics[width=0.975\linewidth]{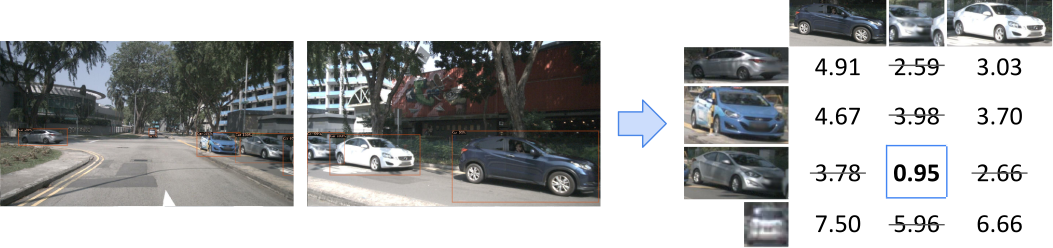}
} \\
\subfloat[Re-identified pairs made in the 6 images, that have been marked as True Positives (TP)]{
\includegraphics[width=0.975\linewidth]{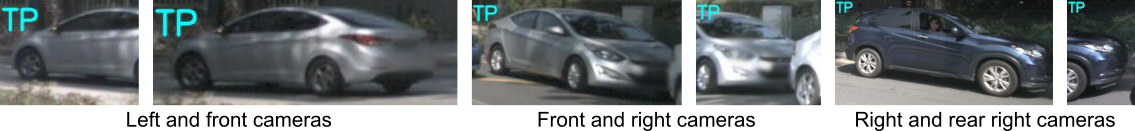}
}
\caption{Re-Identification evaluation process step-by-step}
\label{fig:reid-eval}
\end{figure}

To evaluate the quality of the re-identifications at the image stage, an algorithm has been developed. It performs the following steps: 

\begin{enumerate}
    \item For each group of images at each instant the detections in contiguous images are compared. This distinction is made to speed up the evaluation procedure since there is no possibility of re-identifiable detections between two images that are not adjacent.
    \item For each pair of images, all possible combinations between detections of the same class are compared. In this case, no additional geometric constraints are imposed, since we want to test the performance of the embedding generation module and thus the challenge is more significant. 
    \item A matrix of distances between the comparable detected objects is calculated (having eliminated the pairs of different classes in the previous step) and the pairs whose distance is below a threshold are chosen following the Hungarian method. 
    \item A comparison of whether the obtained pairs actually correspond to the same object according to the ground truth is made and the statistics of TP, TN, FP, and FN are elaborated. 
\end{enumerate}
A visual example of this process can be seen in Figure~\ref{fig:reid-eval}.

 \section{EXPERIMENTAL RESULTS}
 \label{sec:experiments}
For evaluation, the nuScenes dataset has been used, as it provides suitable sensor configuration and annotations for assessing the performance of the presented approach: in the 2D space, to validate the re-identification capabilities of the networks and its effect on the baseline model; and in the 3D space, so that it can be compared to traditional alternatives. To accommodate the original label for the 2D experiments, the minimum bounding boxes have been obtained from the projection of the 3D boxes in each of the cameras. Moreover, the unique ID information of each obstacle has been kept to allow subsequent re-identification validation. 
 
For the experiments of the first part of the configuration, those concerning the 2D detection in the image, the metrics of the KITTI Object \cite{kitti_object} dataset will be used. The minimum overlap percentage between detections and annotations required to consider an object as detected is 50\% for all classes and the difficulty level selected is Moderate (Min. bounding box height: 25 Px, Max. occlusion level: Partly occluded, Max. truncation: 30\%). Although only the 3 most commonly used classes (Pedestrian, Car, and Cyclist) are shown in the tables, the evaluation is performed for all the classes of the nuScenes dataset, and the \textit{all} column is calculated with the weighted average of all of them. The reID columns have been obtained following the procedure explained in the previous Section~\ref{sec:reid-eval}.

In order to evaluate the performance of the whole pipeline after the integration of the re-identification module, the official nuScenes Detection Benchmark metrics are used.

\subsection{Ablation Studies. Number of Embedding Dimensions}
First, an analysis of the adjustment of the hyperparameters of the network has been made, choosing the original configuration of the siaNMS network \cite{cortes2020sianms} but varying the number of dimensions of the output. Results are obtained for the network detection output and the re-identification branch for each of the output dimensions. In addition, we compare the results with those obtained in an equivalent training but without the re-identification branch included. 
This experiment has been performed for two main network configurations: (1) the first one with 2000 pre-NMS proposals and 1000 post-NMS proposals in training, 1000 pre- and post-NMS proposals in test, shown in Table~\ref{tab:num-dim-1000}; and (2) the second with 1000 pre-NMS proposals and 500 post-NMS proposals in training, 500 pre- and post-NMS proposals in the test stage, shown in Table~\ref{tab:num-dim-500}. Both configurations have been tested following the KITTI object image detection metrics for all nuScenes classes (Pedestrian, Car, Cycle, Cyclist, Bus, Truck, Construction Vehicle, Trailer, Barrier, and Cone) for the three frontal cameras of the validation split of the nuScenes dataset.
The \emph{Cycle} and \emph{Cyclist} classes have been obtained by selecting the nuScenes \emph{Bicycle} and \emph{Motorcycle} classes and combining them with the \textit{with\_rider} attribute. This way the \emph{Bicycle} and \emph{Motorcycle} objects with rider are \emph{Cyclists}, and the ones without rider are \emph{Cycles}.

 \begin{table}[ht!]
\centering
\begin{tabular}{@{}cccccccc@{}}
\toprule
\multirow{2}{*}{num dims} & \multicolumn{3}{c}{Classes (AP [\%])} & & \multicolumn{3}{c}{reID} \\ \cmidrule(l){2-5} \cmidrule(l){6-8} 
 & Ped & Car & Cyclist & all & precision & recall & f-score \\ \midrule
off & 42,56 & 51,03 & 38,01 & 40,857 & - & - & - \\
10 & 47,41 & 56,87 & 39,28 & 45,076 & 0,835 & 0,847 & 0,841 \\
15 & \underline{\textbf{52,23}} & \textbf{60,80} & \textbf{41,50} & \textbf{47,827} & 0,861 & 0,838 & 0,850 \\
20 & \textbf{51,17} & 60,53 & 41,15 & 47,381 & 0,864 & 0,849 & 0,857 \\
25 & 45,20 & 51,39 & 34,87 & 41,282 & 0,859 & 0,853 & 0,856 \\
30 & 41,79 & 53,97 & 35,16 & 41,872 & 0,863 & \textbf{0,854} & 0,858 \\
40 & 50,96 & \underline{\textbf{63,12}} & \underline{\textbf{42,09}} & \underline{\textbf{48,501}} & 0,870 & 0,852 & \textbf{0,861} \\
50 & 42,90 & 51,58 & 36,33 & 41,161 & \textbf{0,872} & 0,845 & 0,858 \\ \bottomrule
\end{tabular}%
\caption{Evaluation results for the proposed approach while varying the output layer dimension of the siaNMS branch. 1000 proposals are selected during inference.  The best result in each column is marked in bold and underlined and the second best result is in bold only.}
\label{tab:num-dim-1000}
\vspace{\floatsep}
\centering
\begin{tabular}{@{}crrrcrrr@{}}
\toprule
\multirow{2}{*}{num dims} & \multicolumn{3}{c}{Classes (AP [\%])}&& \multicolumn{3}{c}{reID}\\ \cmidrule(l){2-5} \cmidrule(l){6-8} 
& \multicolumn{1}{c}{Ped} & \multicolumn{1}{c}{Car} & \multicolumn{1}{c}{Cyclist} & med& \multicolumn{1}{c}{precision} & \multicolumn{1}{c}{recall} & \multicolumn{1}{c}{f-score} \\ \midrule
off  & 41,56 & 49,05 & 31,77 & 38,896 & \multicolumn{1}{c}{-} & \multicolumn{1}{c}{-} & \multicolumn{1}{c}{-} \\
10 & 41,04 & 48,90 & 29,51 & 38,478 & 0,856 & 0,852 & 0,854 \\
15& 44,22 & 51,33 & \underline{\textbf{39,96}} & 40,408 & 0,859 & 0,849 & 0,854 \\
20 & 43,74 & 49,20 & 29,93 & 39,666 & 0,868 & 0,850 & 0,859 \\
25 & 43,77 & 50,96 & 31,71 & 40,366 & \textbf{0,873} & \textbf{0,874} & \textbf{0,874} \\
30 & \underline{\textbf{48,69}} & \textbf{53,23} & 38,75 & \textbf{42,612} & 0,869 & 0,872 & 0,870 \\
40 & 43,86 & 50,91 & 37,62 & 40,211 & 0,872 & \underline{\textbf{0,876}} & \textbf{0,874} \\
50 & \textbf{47,09} & \underline{\textbf{54,48}} & \textbf{39,28} & \underline{\textbf{42,699}} & \underline{\textbf{0,877}} & 0,855 & 0,866 \\ \bottomrule
\end{tabular}%
\caption{Evaluation results for the proposed approach while varying the output layer dimension of the siaNMS branch. 500 proposals are selected during inference.  The best result in each column is marked in bold and underlined and the second best result is in bold only.}
\label{tab:num-dim-500}
\end{table}

As can be seen in Tables~\ref{tab:num-dim-1000}~and~\ref{tab:num-dim-500}, the results of both the re-identification branch and the detection and classification branches vary depending on the output size of the re-ID branch. This is because the training of the network is done in an end-to-end manner, and the ability of the network to generalize is altered by the introduction of the third branch. That said, it can be generally observed that the inclusion of the obstacle re-identification branch improves the network's ability to detect and classify obstacles, which implies that the quality of the feature maps obtained in the intermediate layers of the network is improved by its optimization for a complementary task, as previously observed in the literature \cite{he2017mask, liang2019multi}. It is worth highlighting the results marked in bold and underlined, where an improvement of more than 10\% is achieved for some classes, improving the performance of the network for all classes by almost 8\%. Likewise, we see that the re-identification results for all cases are similar and relatively good.

\subsection{Ablation Studies. Re-Identification Branch Configuration}
In the next experiment, we study the effect of changes in the structure of the re-identification branch. For this purpose, several alternatives have been proposed, with different numbers of neurons and intermediate layers, or eliminating altogether the Convolutional layers and keeping only the Fully Connected ones, in a similar way as the other two output branches are constructed. Table~\ref{tab:abl_network_layers} shows the configuration of all the studied options. Finally, Table~\ref{tab:results_layers} shows the detection and re-identification results for each design. For all these configurations a constant number of output dimensions, 25, has been chosen so that the results are comparable. This number of output dimensions has been decided based on the configuration that gave the best results in the re-ID evaluation (see Table~\ref{tab:num-dim-500}). For the following experiment, 500 proposals during the test phase have been chosen. Once more, the evaluation has been performed following the KITTI object image detection metrics for all nuScenes classes, but this time the six cameras of the validation split of the nuScenes dataset have been used. 

\begin{table}[!ht]
\centering
\resizebox{\textwidth}{!}{%
\begin{tabular}{@{}lcccccccccccccccc@{}}
\toprule
Conf & \#0 & \#1 & \#2 & \#3 & \#4 & \#5 & \#6 & \#7 & \#8 & \#9 & \#10 & \#11 & \#12 & \#13 & \#14 & \#15 \\
\midrule
k & 3 & 3 & 5 & 5 & 5 & 5 & 3 & 3 & 3 & 3 & 3 & - & - & - & - & - \\
C1 & 32 & 16 & 32 & 16 & 64 & 32 & 32 & 32 & 32 & 32 & 16 & - & - & - & - & - \\
C2 & 64 & 32 & 64 & 32 & 128 & 64 & 64 & 64 & 64 & 64 & - & - & - & - & - & - \\
C3 & - & - & - & - & - & - & - & 64 & 128 & - & - & - & - & - & - & - \\
C4 & - & - & - & - & - & - & - & 128 & 256 & - & - & - & - & - & - & - \\
FC1 & 1024 & 4096 & 1024 & 4096 & 512 & 4096 & 4096 & 1024 & 1024 & 512 & 512 & 1024 & 1024 & 1024 & 1024 & 1024 \\
FC2 & 256 & 1024 & 256 & 1024 & 256 & 1024 & 1024 & 256 & 256 & 256 & 256 & 1024 & 1024 & 512 & 512 & 256 \\
FC3 & 256 & 256 & 256 & 256 & 256 & 256 & 256 & 256 & 256 & 128 & 128 & 256 & - & 128 & - & - \\
\bottomrule
\end{tabular}%
}
\caption{Specifications of the tested configurations for the re-identification branch. k is the kernel size used, C1-C4 is the number of output neurons of the Convolutional Layer, FC1-FC3 is the number of output neurons of the Fully Connected Layer.}
\label{tab:abl_network_layers}
\vspace{\floatsep}
\end{table}

\begin{table}[ht]
\centering
\begin{tabular}{@{}lcccccccc@{}}
\toprule
\multirow{2}{*}{Conf.} & \multicolumn{4}{c}{Classes (AP [\%])}& \multicolumn{3}{c}{reID}\\ \cmidrule(l){2-5} \cmidrule(l){6-8} 
& \multicolumn{1}{c}{Ped} & \multicolumn{1}{c}{Car} & \multicolumn{1}{c}{Cyclist} & all & \multicolumn{1}{c}{precision} & \multicolumn{1}{c}{recall} & \multicolumn{1}{c}{f-score} \\ \midrule
\#0 & 44,42 & 53,39 & 30,84 & 41,253 & 0.817 & 0.846 & 0,832 \\
\#1 &  45,33 & 54,38 & 31,13 & 41,622 & 0,815 & 0,838 & 0,827 \\
\#2 &  42,36 & 53,53 & \textbf{31,64} & 41,048 & 0,816 & 0,844 & 0,830 \\
\#3 &  41,97 & 51,42 & 28,91 & 40,020 & 0,806 & \textbf{0,856} & 0,831 \\
\#4 &  44,23 & 53,18 & 29,81 & 41,209 & 0,836 & 0,841 & 0,839 \\
\#5 &  42,47 & 53,25 & 28,78 & 40,814 & 0,818 & 0,851 & 0,835 \\
\#6 &  44,53 & 53,49 & 31,25 & 41,525 & 0,826 & 0,846 & 0,836 \\
\#7 &  45,83 & 54,5 & 29,43 & 42,261 & 0,728 & 0,829 & 0,779 \\
\#8 &  45,15 & 53,28 & 30,48 & 40,998 & 0,728 & 0,839 & 0,784 \\
\#9 &  45,71 & 51,61 & 29,88 & 40,653 & 0,818 & 0,846 & 0,832 \\
\#10 &  44,43 & 53,63 & 30,66 & 41,358 & 0,817 & 0,855 & 0,836 \\
\#11 &  \textbf{45,90} & \textbf{55,16} & 29,54 & \textbf{42,353} & 0,857 & 0,845 & 0,851 \\
\#12 &  42,52 & 53,1 & 29,01 & 41,016 & \textbf{0,863} & 0,844 & 0,854 \\
\#13 &  44,45 & 54,62 & 28,96 & 41,659 & 0,860 & 0,849 & \textbf{0,855} \\
\#14 &  42,13 & 51,68 & 29,07 & 40,278 & 0,851 & 0,848 & 0,850 \\
\#15 &  41,98 & 51,48 & 31,19 & 40,294 & 0,859 & 0,836 & 0,848 \\
\bottomrule
\end{tabular}%
\caption{Results of the evaluation of the different configurations for the re-identification branch within the image detection network. Details of each conf. are given in Table~\ref{tab:abl_network_layers}. The best result for each column is highlighted in bold.}
\label{tab:results_layers}
\end{table}

Based on the reference configuration \#0, as shown in Table~\ref{tab:results_layers}, variations made in the other configurations have a slight effect on the obtained results. 
Although the alterations in the network composition yield minor performance changes, configuration \#11 presents the best overall results, achieving the highest precision for pedestrians (45.9\%), cars (55.16\%), and all classes combined (42.353\%), and a re-ID f-score (0.851) in pair with the best case scenario.
 
\subsection{Image Detection Qualitative Results}
To visually assess the operation of the pipeline, some qualitative results in the nuScenes benchmark are presented. The model used follows the configuration \#11 from \ref{tab:abl_network_layers} with an output dimension of 30, using 500 proposals in the RPN stage. For the remaining experiments, these hyperparameters are kept.

\begin{figure}[ht]
\captionsetup[subfloat]{justification=centering}
\centering
\subfloat[]{
\centering
\includegraphics[width=0.975\textwidth]{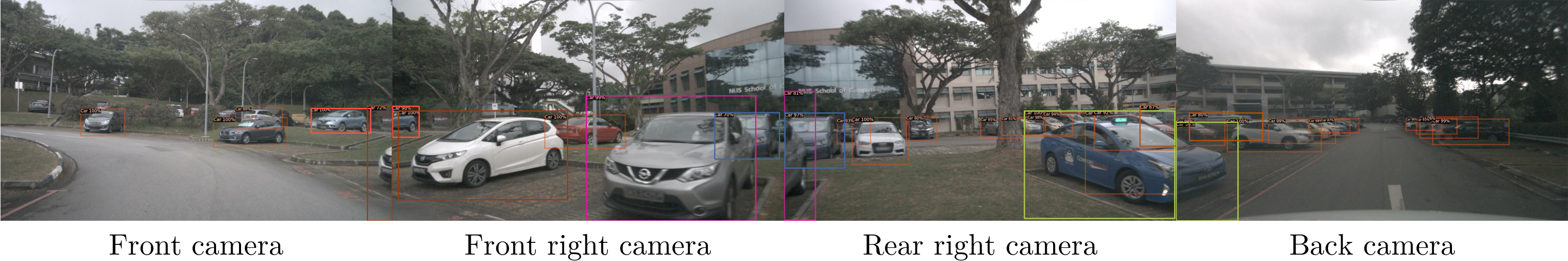}
} \\ 
\subfloat[]{
\centering
\includegraphics[width=0.975\textwidth]{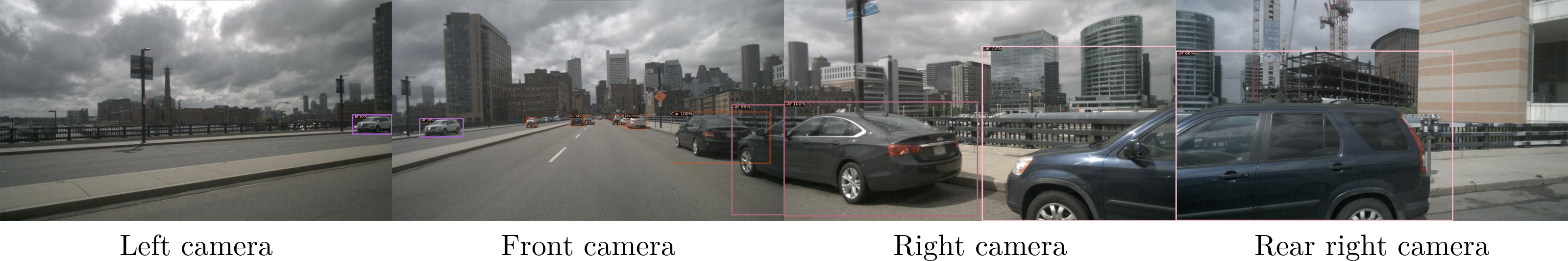}
} \\
\subfloat[]{
\centering
\includegraphics[width=0.975\textwidth]{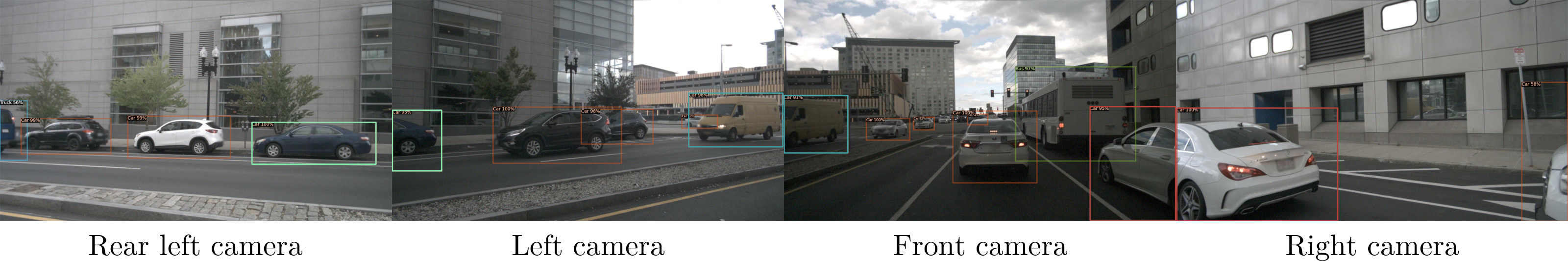}
}
\caption{Qualitative examples of the image detections. The same color bounding box indicates that the two obstacles from different cameras have been re-identified as the same. For the rest of the bounding boxes: orange for cars, blue for pedestrians, and purple for cyclists.}
\label{fig:qualitative}
\end{figure}

Some examples of detection and re-identification results are shown in Figure~\ref{fig:qualitative}. The images show frames captured by the different cameras at the same temporal instant. It can be seen how obstacle detection even at long distances is appropriately done and how the obstacles appearing in adjacent cameras are re-identified (the bounding box of the re-identified obstacles is drawn with the same color, chosen randomly).

\subsection{3D Obstacle Detection Quantitative Results}

In this section, we evaluate the second stage of the proposed detection pipeline, which involves taking as input a filtered point cloud in the form of a frustum for each of the objects detected in the previous image-based detection stage and processing them through a series of pointnet-based networks to estimate the parameters defining a three-dimensional bounding box -position ($x, y, z$), dimensions ($l, w, h$) and orientation ($\theta$)- that encapsulates the detected obstacle in the image space. 

As introduced In \cite{cortes2020sianms}, to leverage the outcome of the re-identification step in the 3D box estimation, the \textit{region2frustum} operation shown in \ref{fig:global_graph} needs to be adapted so that it can produce more refined filtered point cloud candidates from matched detections across adjacent cameras:
\begin{enumerate}
    \item For detections appearing in a single camera, we follow the process outlined in the original F-PointNets paper. A projection of the point cloud onto the bounding box of the obstacle is performed, and the points within the box are selected.
    \item For detections appearing in two cameras simultaneously, we employ a union of the filtered point clouds from each camera. The process involves the following steps:
    We first obtain the filtered frustum for each individual detection, following the same procedure as for unique detections.
    Next, we verify if the point clouds obtained for each instance have common points. If there are no shared points, it is considered a false positive re-identification, and the merging of objects is discarded.
    If shared points exist the match is considered spatially coherent, indicating a correct re-identification. The two filtered point clouds are merged, retaining only the unique points.
    Finally, the central axis of the detection is computed by calculating the average angle between the two outermost angles from the detections in both cameras. A deeper explanation of this process can be read in \cite{cortes2020sianms}.
\end{enumerate}

To evaluate the results, the proposed method (\textit{siaNMS}) is studied against three other approaches. The \textit{Original} configuration, namely the results obtained by the baseline F-PointNets model with the 2D detection network unaltered. A second one, which uses the detection network presented in this article, but without carrying out the union of re-identified objects (\textit{2D+embedding}). A third pipeline, where a non-maximum suppression (NMS) algorithm is applied to the results obtained from the Original configuration (\textit{Original+NMS}), such as the one presented in the original paper \cite{cortes2020sianms}. 


The assessment encompasses the different metrics provided by the nuScenes dataset, considering various aspects such as positional error, size error, and orientation error, among others, for all annotated types of obstacles. More details can be found in the nuScenes object detection task \cite{nuscenes}.

Table~\ref{tab:table-carpedcyc} shows the results for Average Precision (AP $\uparrow$)  [\%], Average Translation Error (ATE $\downarrow$) [m], Average Size Error (ASE $\downarrow$) [\%] and Average Orientation Error (AOE $\downarrow$) [rad] for the car, pedestrian, and cyclist classes. Similarly, Table~\ref{tab:table-all} shows the averaged results across all classes. The experiments have been done in the following manner: The Frustum Pointnets detection network has been trained with the frames of the training split of the nuScenes dataset and the validation split of the nuScenes dataset has been used for evaluation. 
To better understand the impact of the proposed approach, the same evaluation has been performed taking into account only the object instances that appear in more than one camera.

\begin{table}[ht]
\resizebox{\textwidth}{!}{%
\centering
\begin{tabular}{lllccccccccccc}
\hline
\multirow{2}{*}{Area}    & \multirow{2}{*}{} & \multicolumn{4}{c}{car}                                         & \multicolumn{4}{c}{pedestrian}                                  & \multicolumn{4}{c}{cyclist}                                     \\ \cline{3-14} 
                         &                   & AP            & ATE            & ASE           & AOE            & AP            & ATE            & ASE           & AOE            & AP            & ATE            & ASE           & AOE            \\ \hline
\multirow{4}{*}{all}     & Original          & 48.2          & 0.434          & 17.1          & 0.493          & 58.4          & 0.263          & 28.5          & 1.194          & 44.2          & 0.252          & 26.1          & 0.898          \\
                         & 2D+embedding      & 47.8          & 0.427          & 17.1          & 0.486          & 57.1          & 0.259          & \textbf{28.4} & 1.200          & 43.9          & 0.247          & 26.4          & 0.890          \\
                         & Original + NMS    & 48.7          & 0.436          & 17.1          & 0.487          & 56.0          & 0.259          & 28.5          & \textbf{1.188} & \textbf{46.6} & 0.257          & \textbf{26.1} & 0.903          \\
                         & siaNMS            & \textbf{51.5} & \textbf{0.422} & 17.1          & \textbf{0.476} & \textbf{59.7} & \textbf{0.246} & 28.5          & 1.203          & 46.3          & \textbf{0.243} & 26.2          & \textbf{0.874} \\ \hline
\multirow{4}{*}{overlap} & Original          & 41.7          & 0.439          & 16.8          & 0.477          & 44.2          & 0.355          & 28.4          & 1.136          & 34.0          & 0.301          & 26.7          & 0.843          \\
                         & 2D+embedding      & 41.6          & 0.433          & \textbf{16.7} & 0.463          & 43.5          & 0.359          & \textbf{28.3} & \textbf{1.133} & 32.5          & 0.288          & 26.4          & 0.799          \\
                         & Original + NMS    & 44.8          & 0.442          & 16.8          & 0.472          & 47.9          & 0.345          & 28.4          & 1.138          & 35.7          & 0.311          & 26.3          & 0.876          \\
                         & siaNMS            & \textbf{50.1} & \textbf{0.416} & \textbf{16.7} & \textbf{0.435} & \textbf{50.2} & \textbf{0.330} & 28.6          & 1.148          & \textbf{41.5} & \textbf{0.261} & \textbf{26.0} & \textbf{0.597} \\ \cline{1-14} 
\end{tabular}
}
\caption{Comparison of the 3D Object Detection Performance on the nuScenes Validation Set in Different Regions of Interest. Results for Car, Pedestrian, and Cyclist Classes. The best result for each column is highlighted in bold.}
\label{tab:table-carpedcyc}
\end{table}

\begin{table}[ht]
\centering
\small
\begin{tabular}{llcccc}
\hline
Area                     &                & AP             & ATE            & ASE            & AOE            \\ \hline
\multirow{4}{*}{all}     & Original       & 32.87          & 0.526          & 29.81          & 0.928          \\
                         & 2D+embedding   & 31.87          & 0.525          & 29.72          & 0.930          \\
                         & Original + NMS & 32.46          & 0.529          & 29.75          & 0.925          \\
                         & siaNMS         & \textbf{33.18} & \textbf{0.521} & \textbf{29.58} & \textbf{0.915} \\ \hline
\multirow{4}{*}{overlap} & Original       & 25.59          & 0.585          & 30.15          & 0.947          \\
                         & 2D+embedding   & 24.93          & 0.588          & 29.93          & 0.929          \\
                         & Original + NMS & 26.90          & 0.583          & 29.99          & 0.954          \\
                         & siaNMS         & \textbf{28.40} & \textbf{0.573} & \textbf{29.80} & \textbf{0.899} \\ \cline{1-6} 
\end{tabular}%
\caption{Comparison of the 3D Object Detection Performance on the nuScenes Validation Set in Different Regions of Interest. Results for All Classes. The best result for each column is highlighted in bold.}
\label{tab:table-all}
\end{table}

As can be seen in Tables~\ref{tab:table-carpedcyc}~and~\ref{tab:table-all}, the AP results of all classes improve when using the method proposed in the article in contrast to conventional methods. 
For the car class, the average accuracy increases by more than 3\% with respect to the original method, more than 1\% for pedestrians, and more than 2\% for cyclists. Although these numbers may not suggest a major effect of the proposed additional re-identification branch, its real impact is being diluted by the fact that only a reduced portion of objects leads to redundant detections and is affected by truncation.

As a consequence, when the results are analyzed considering only those objects falling in the regions of overlap between contiguous cameras, it can be observed that these improvements are much more accentuated, especially in bulky categories such as cars -more prone to be seen from multiple views- where the gain in AP goes up to 8.4\%. 
In addition to improving the number of detections, the quality of the detections is also affected, reducing the average errors of position, size, and orientation of all classes with respect to the original method and practically all with respect to a conventional NMS method.

\subsection{3D Object Detection Qualitative Results}
\begin{figure}
\captionsetup[subfloat]{justification=centering}
\subfloat[]{
    \includegraphics[height=7.8cm]{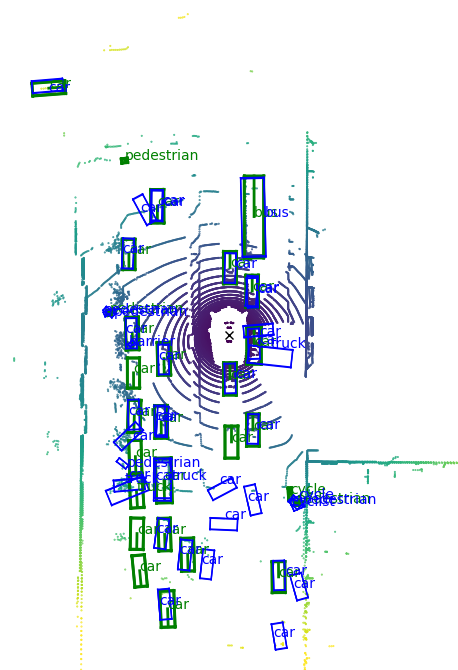}\\ \vspace{1em}
    \includegraphics[height=7.8cm]{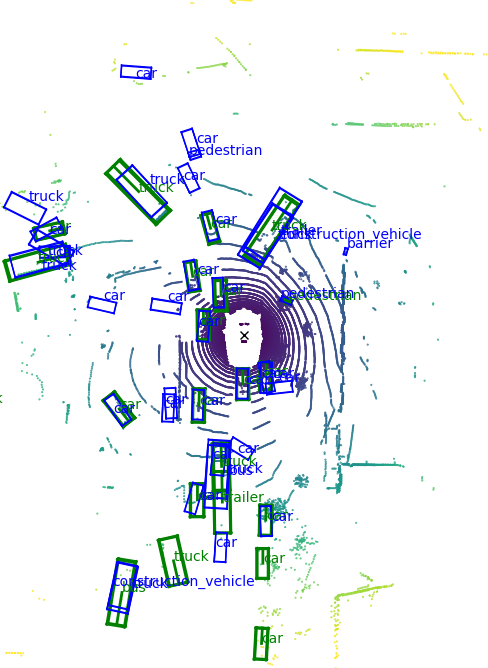}\\ \vspace{1em}
    \includegraphics[height=7.8cm]{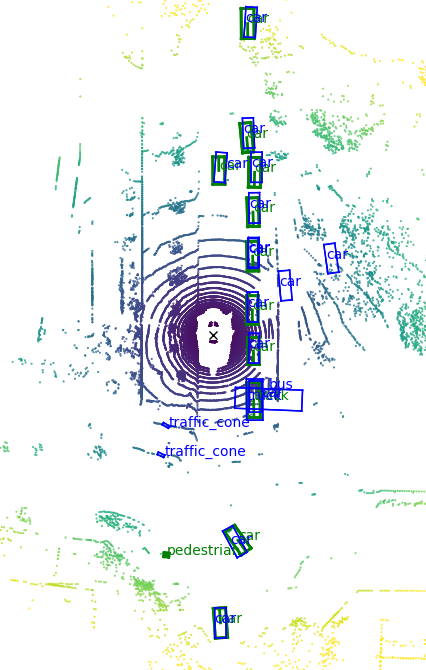}
}\\
\subfloat[]{
    \includegraphics[height=7.8cm]{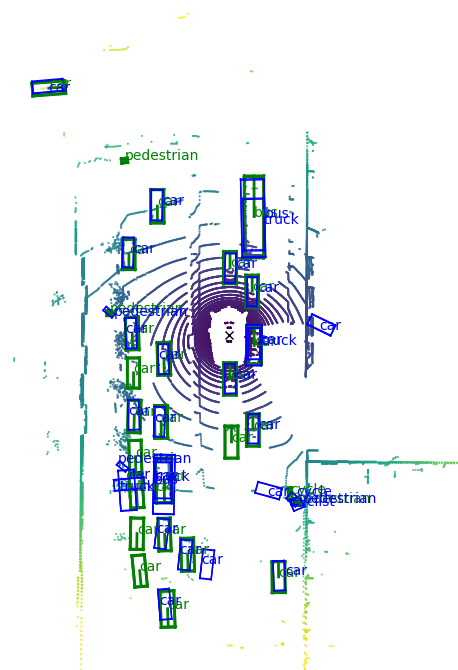}\\ \vspace{1em}
    \includegraphics[height=7.8cm]{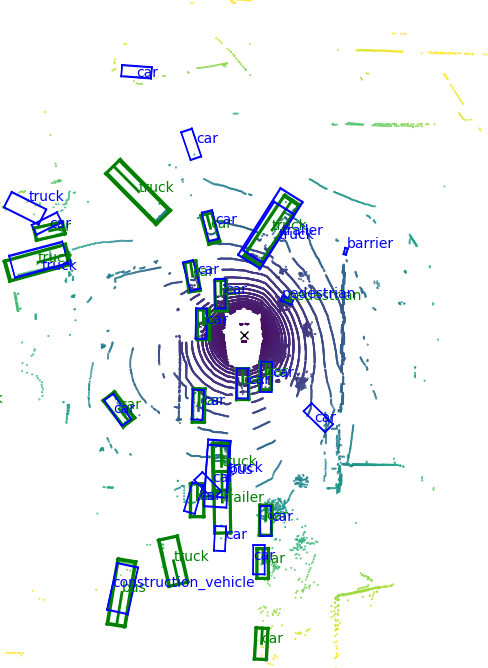}\\ \vspace{1em}
    \includegraphics[height=7.8cm]{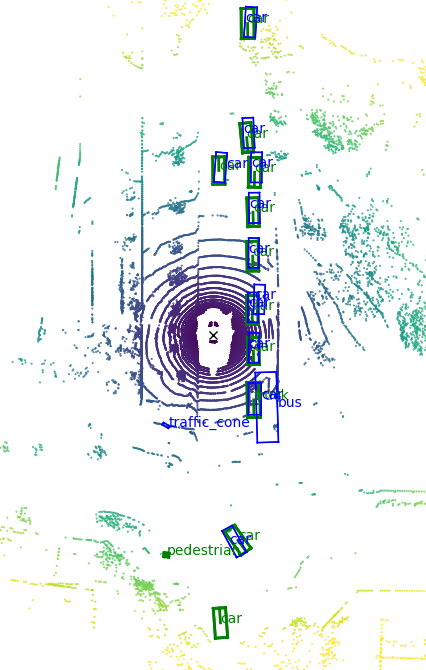}
}
\caption{Results on nuScenes validation set. In (a) the original approach with NMS is used, while in (b), the proposed architecture is used.}
\label{fig:examples}
\end{figure}
Similar to what was done to visualize the impact of the proposal in the image space, this section includes qualitative 3D results of the final system compared to the ones of the reference framework.
Here the detections are presented in the LiDAR bird's-eye view, which displays the LiDAR readings as seen from an orthographic top view, where each cell represents a square pillar lying in a theoretical ground plane. The images in Figure~\ref{fig:examples} show some frames from the validation split of the nuScenes dataset comparing the two best approaches (i.e. \textit{Original+NMS} and \textit{siaNMS}). In green are the ground truth boxes, while the boxes obtained by the full pipeline proposed in this article are painted in blue. As can be seen in the three cases, the number of false positives is reduced, and the quality of the detections is improved in terms of positioning and orientation of the obtained boxes.

\section{CONCLUSIONS}
\label{sec:conclusion}
In this work, the capabilities of a well-established camera-LiDAR sequential fusion 3D object detection pipeline have been enriched to make it more suitable for its use in multi-camera setups, where its performance deteriorates when duplicate detections of the same object appear.

To this end, the 2D object detector module of its first stage has been extended by means of an embedding branch that enables the re-identification of instances across contiguous cameras. Unlike the predecessor study, the siaNMS block is now integrated as part of the decoding layers of the network, in a multi-task fashion, allowing for its training in an end-to-end manner.

The presented approach has led to an improvement in the quality of the features extracted by the network, boosting not only the re-ID outcomes but also the primary function of the network, the detection, and classification of obstacles in the image, as demonstrated in the conducted analysis on the nuScenes dataset.

Further experimentation shows that the early elimination of redundant detections also benefits the results provided by the 3D box characterization model, reducing the negative effects of truncated boxes on the image plane and thus contributing to a significant gain in the overall performance of the system.



\bibliographystyle{unsrt}  
\bibliography{paper}

\end{document}